%% file: main.tex
\documentclass[letterpaper, 10 pt, conference]{ieeeconf}  

\IEEEoverridecommandlockouts                              

\usepackage{graphics} 
\usepackage{epsfig} 
\usepackage{mathptmx} 
\usepackage{times} 
\usepackage{amsmath} 
\usepackage{amssymb}  
\usepackage[T1]{fontenc}

\usepackage{multirow} 
\usepackage{wrapfig} 
\usepackage{caption}
\usepackage{subcaption}
\usepackage{amssymb}
\usepackage{xcolor}
\usepackage{bbding}
\usepackage{pifont}
\usepackage{wasysym}
\usepackage{amssymb}
\usepackage{mathtools}
\usepackage{bm}
\renewcommand{\vec}[1]{\bm{#1}}

\newcommand{\set}[1]{\bm{#1}}
\newcommand{\realnum}[0]{\mathbb{R}}
\newcommand{\accro}[0]{VIRDO}
\usepackage[utf8]{inputenc}
\usepackage[english]{babel}
\usepackage[
backend=biber,
style=ieee,
citestyle=ieee
]{biblatex}
\title{\LARGE \bf
VIRDO: Visio-tactile Implicit Representations of Deformable Objects
}

\author{Youngsun Wi$^{1}$, Pete Florence$^{2}$, Andy Zeng$^{2}$ and Nima Fazeli$^{1}$
\thanks{$^{1}$ Youngsun Wi and Nima Fazeli are with the Robotics Department at the University of Michigan, MI, USA 
{\tt\small <yswi,nfz>@umich.edu}}%
\thanks{$^{2}$ Pete Florence and Andy Zeng with Robotics at Google, Mountain View, CA, USA
{\tt\small <peteflorence,andyzeng>@google.com}}%
}

\begin{document}

\maketitle
\thispagestyle{empty}
\pagestyle{empty}

\begin{abstract}
    Deformable object manipulation requires computationally efficient representations that are compatible with robotic sensing modalities. In this paper, we present \accro: an implicit, multi-modal, and continuous representation for deformable-elastic objects.
    \accro~operates directly on visual (point cloud) and tactile (reaction forces) modalities and learns rich latent embeddings of contact locations and forces to predict object deformations subject to external contacts.
    Here, we demonstrate \accro s~ability to: i) produce high-fidelity cross-modal reconstructions with dense unsupervised correspondences, ii) generalize to unseen contact formations, and iii) state-estimation with partial visio-tactile feedback.
    
    \url{https://github.com/MMintLab/VIRDO}
\end{abstract}



\input{text/01-introduction}


\input{text/03-methodology}


\input{text/04-results}


\input{text/02-related-work}

\input{text/05-discussion}

\section{Acknowledgement}
This research is partly supported by Robotics at Google.

\printbibliography
\end{document}

%% file: text/01-introduction.tex
\section{Introduction}

\begin{figure}[thpb]
  \centering
    \includegraphics[width=0.5\textwidth]{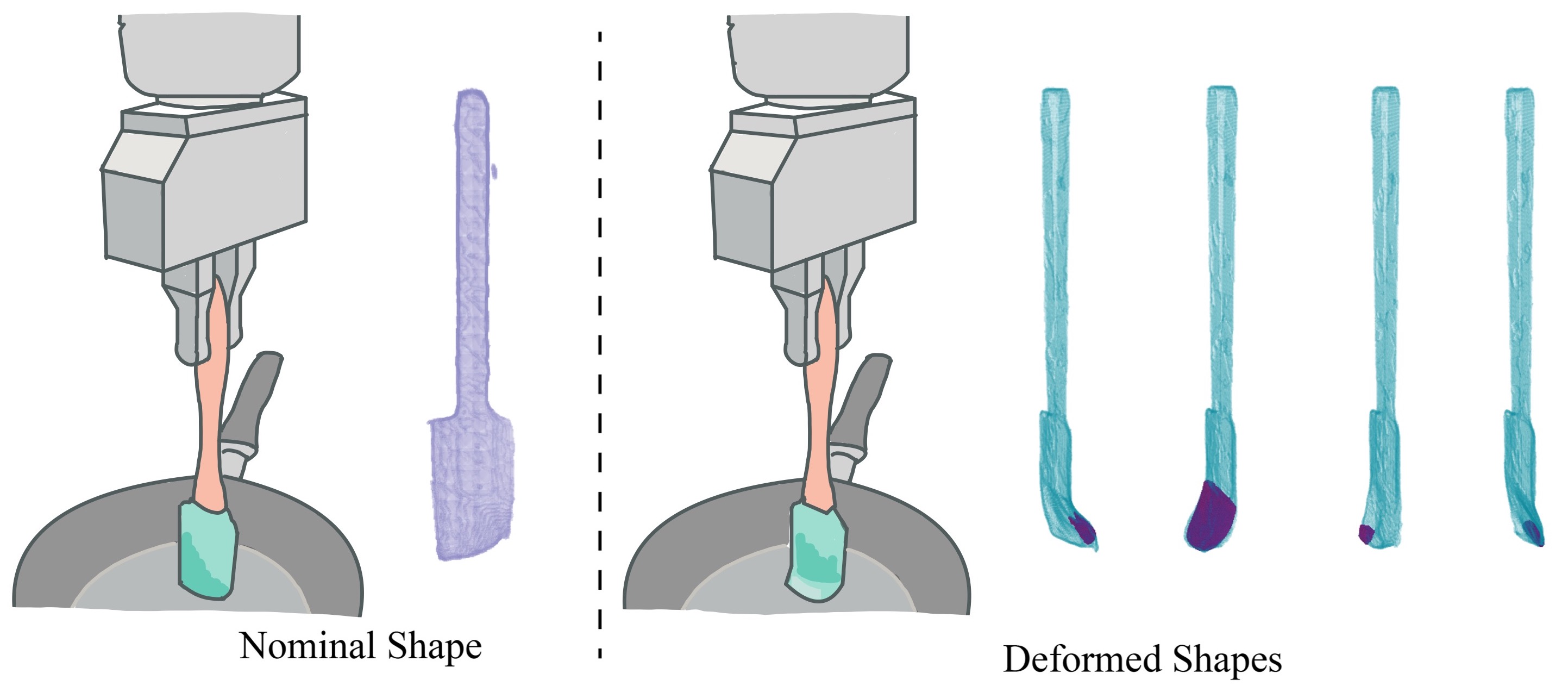}
    \caption{\accro~represents objects and their deformations due to external contact forces as signed distance functions (SDF) via decoding of latent shape and force codes using feed forward neural networks. The external contacts on the deformed objects are shaded in purple.}
    \vspace{-15pt}
  \label{fig:SDF}
\end{figure}

Dexterous manipulation of deformable objects is an important open problem in robotics~\cite{billard2019trends,yin2021modeling}. These objects are ubiquitous in our day-to-day lives and play a key role in many applications including manufacturing, in-home assistive care, surgery, and cooking, as shown in Fig.~\ref{fig:SDF}. Despite their prevalence and importance, deformable objects have received less attention than their rigid counterparts owing to their inherent complexities in modeling, perception, and controls \cite{yin2021modeling,sanchez2018robotic,arriola2020modeling,hou2019review}. To illustrate, the states of rigid bodies with known geometries can be succinctly represented with 6D pose and velocity. However, deformable objects have an infinite continuum of states and their representation and perception remains an open problem \cite{yin2021modeling}.

In this paper, we present VIRDO -- an implicit, dense, cross modal, and continuous architecture that addresses these fundamental representation and perception challenges for the class of elastically deformable objects. The central feature of our method is learning deformation fields informed by cross modal visual and tactile cues of external contacts. We further contribute a dataset of elastically deformable objects with boundary conditions used to evaluate. This paper focuses on dense geometric representations because they can facilitate downstream tasks such as state-estimation from partial views and estimating dense correspondences, as we demonstrate, as well as bootstrapping keypoint/affordance learning.

Our method draws inspiration from recent theoretical and computational advances in continuous neural representations parameterized by feed forward neural networks that have gained traction in computer vision and graphics \cite{pratt2017fcnn,park2019deepsdf,sitzmann2020implicit}. These networks are trained to map low-dimensional inputs (e.g. $\realnum^3$ spatial queries) to output representations of shape or density at each input location~\cite{tancik2020fourier}. These networks have been used to represent signed distances \cite{pratt2017fcnn,park2019deepsdf,sitzmann2020implicit}, occupancy \cite{mescheder2019occupancy}, and volume density \cite{mildenhall2020nerf} where state-of-the-art result have been demonstrated across a variety of shape representation tasks \cite{chen2019learning,deng2019nasa,genova2019learning,genova2020local,michalkiewicz2019implicit,yen2020inerf}. This paper takes steps towards the successful adoption of these approaches for robotic applications by integrating tactile sensing with vision.

\subsection{Problem Statement \& Assumptions}
\vspace{-5pt}

Our goal is to derive a computationally efficient and generative model that: 1) predicts object deformations subject to external forces; and 2) is compatible with common robotic sensors. We assume the object geometry is described by its point cloud: an unordered set $\set{P}:=\{ \vec{p} \in \realnum^{3} : \text{SDF}(\vec{p}) = 0\}$ where SDF denotes the signed distance w.r.t. the surface of the object. Point clouds are obtained from commodity depth sensors or 3D scanners commonly found in industry. Contact locations are also given as a set of points $\set{Q}$ which can be given by an upstream perception algorithm such as \cite{hermans2013learning,sharma2020relational,le2010learning}. For the tactile input, net reaction force is given by $\vec{u} \in \realnum^3$ at the wrist of the robot which can be measured by common industrial F/T sensors or recovered from joint torques. In this paper, we derive a continuous and implicit representation of the deformed object geometry ($f(\set{P}, \set{Q}, \set{u})=s$) subject to external forces and their locations. Here the object geometry is given by the zero-level set of the implicit function; i.e. $s=0$. Fig.~\ref{fig:arch} depicts an expected setup and our representation, \accro.

%% file: text/03-methodology.tex
\section{Representing Deformable Objects using SDFs}
\begin{figure*}[thpb]
    \centering
    \includegraphics[width=\textwidth]{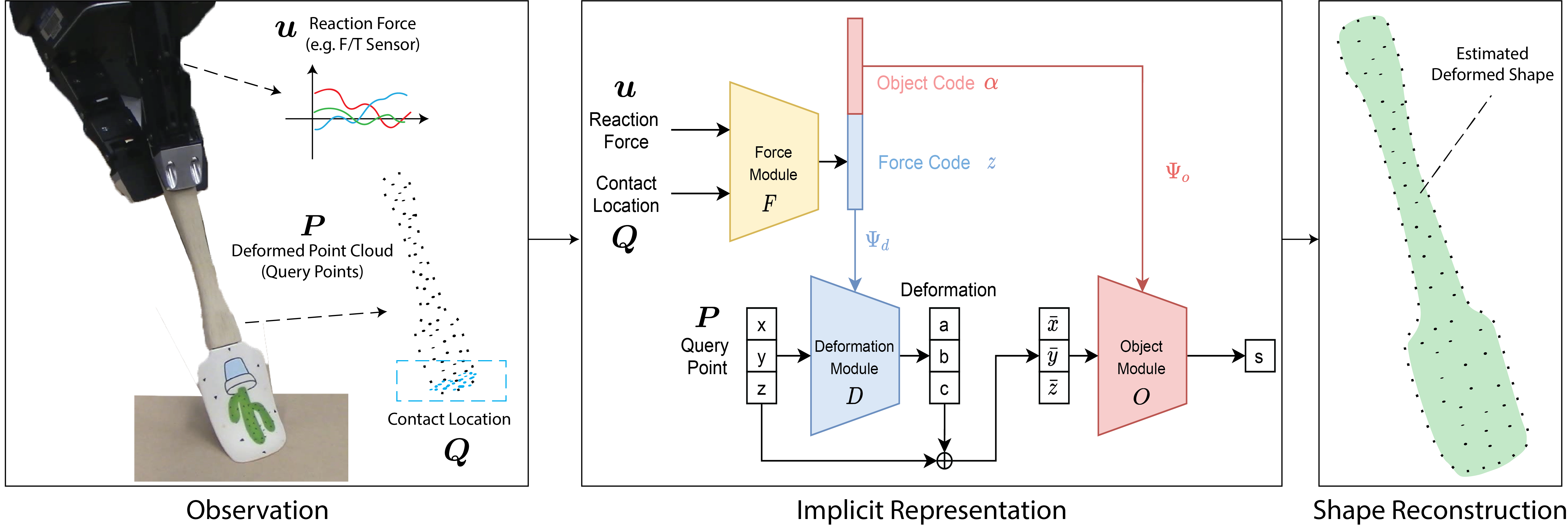}
    \caption{\textbf{Representation Architecture:} The left panel depicts how visual data in the form of point clouds and tactile in the form of reaction forces may be collected in practical robotic settings. The middle panel depicts the network and how this information is processed to predict the implicit surface representation encoded as a signed-distance function. Finally, the right panel depicts the reconstruction of the estimated true surface given the external contacts and reaction force.}
    \label{fig:arch}
\vspace{-4mm}
\end{figure*}

At a high-level, \accro~decomposes object representations into a nominal shape representation and a point-wise deformation field. Here, we choose signed distance fields as our underlying representation and discuss this choice in Sec.~\ref{sec:discussion}. The nominal shape representation decodes latent shape embeddings $\vec{\alpha} \in \realnum^l$ into continuous signed-distance fields, the zero-level set of which is the undeformed object geometry -- similar to the architectures proposed in \cite{park2019deepsdf,sitzmann2020implicit,tancik2020fourier}. The point-wise deformation field is produced using a summary of all boundary conditions (contact locations, reaction force, and fixed constraints) leveraging a permutation invariant set operator. The structure is fully differentiable and can be learned end-to-end. In the following, we discuss each component in more detail.
\vspace{-15pt}
\subsection{Nominal Shape Representation} \label{sec:nom}
\vspace{-5pt}

The nominal shape of an object is the geometry it takes in the absence of external contact forces. The nominal geometry is produced by the object module as $\set{O}(\vec{x}|\set\Psi_o(\vec{\alpha})) = s$ as a signed distance field, where $\vec{x} = (x, y, z)$ is a query point, and $s$ is the signed-distance.


\begin{wrapfigure}{r}{0.25\textwidth}
  \begin{center}
    \includegraphics[width=0.25\textwidth]{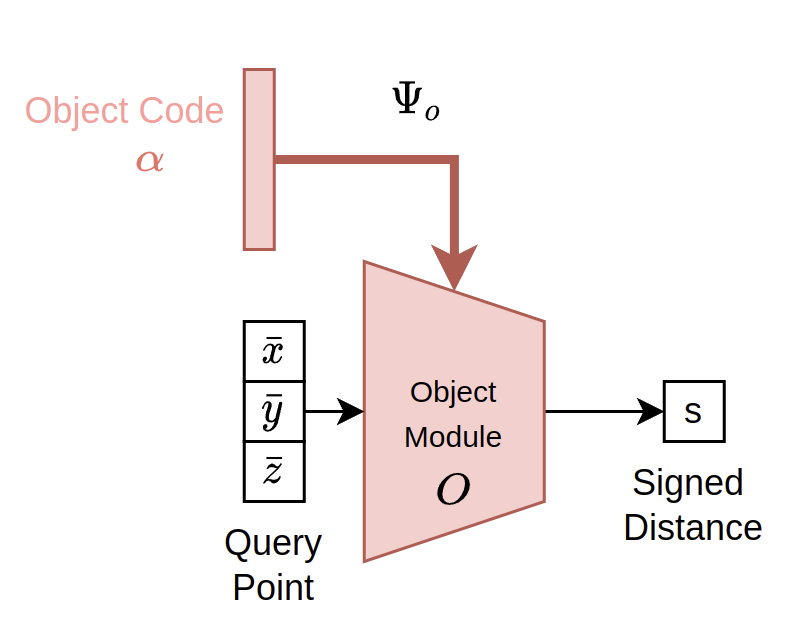}
  \end{center}
  \caption{Object Module Architecture}
  \label{fig:object module}
\end{wrapfigure}

\begin{figure*}[t!]
    \centering
    \includegraphics[width=0.9\textwidth]{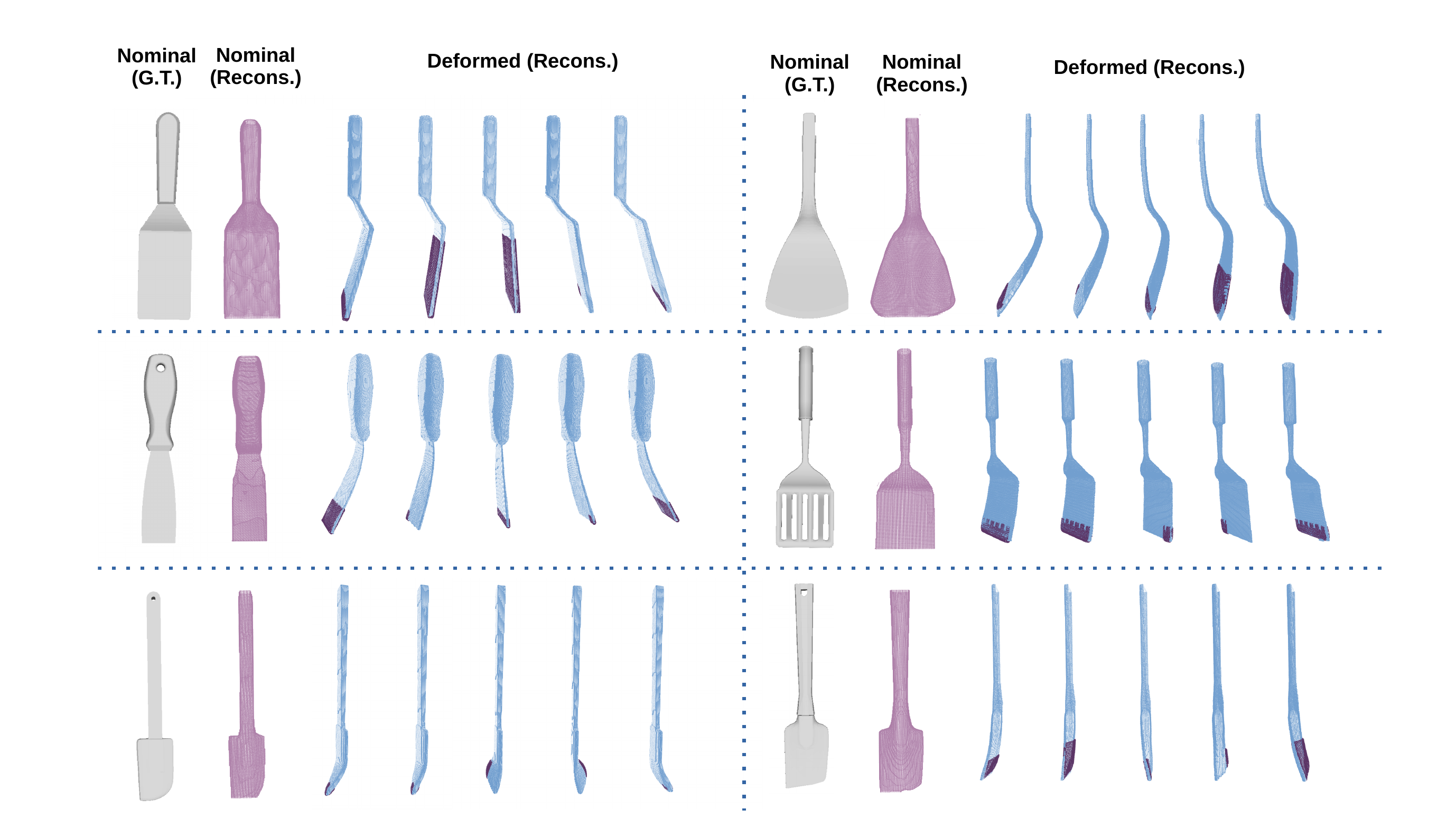}
    \caption{\textbf{Reconstruction Results:} Reconstructions of multiple nominal shapes and their deformations, learned simultaneously by \accro. Marching Cube algorithm is used for the reconstruction, where we highlighted the contact location as purple region.} \label{fig:deformations}
\vspace{-4mm}
\end{figure*}

The purpose of the object code and module is to allow \accro~to represent multiple objects. Here, we use a hyper-network $\set\Psi_{o}(\vec{\alpha})$ to decode the object code into object module parameter $\vec{\theta}_{o}$, similar to \cite{deng2020deformed}. 
Fig.~\ref{fig:deformations} shows example reconstructions of nominal shapes.

\textbf{Pre-training to Learn Nominal Shapes:} We pre-train \accro~on nominal shapes before training on deformations. During this stage, we initialize the object codes as $\vec\alpha \sim N(0,0.1^2)$ and update them with hyper-network $\set\Psi_{o}$. The loss function for pretraining nominal shapes is:
\begin{align} \label{eq:train_loss}
    L_{nominal} = L_{sdf} + \lambda_2 L_{latent} + \lambda_3 L_{hyper}.
\end{align}
We introduce the shorthand $\set{O}^i(\vec{x}) = \set{O}(\vec{x} | {\Psi_o}(\vec{\alpha_i}))$ to denote the signed-distance field corresponding to the $i^{th}$ object code at query point $\vec{x}$ and define $L_{sdf}$ as: 
\begin{align*}
L_{sdf} =\sum_{i=1}^N \Big(
& \sum_{\bar{\vec{x}} \in \set{\Omega}} |clamp(\set{O}^i(\vec{x}), \delta)- clamp(s^*,\delta)| \nonumber \\
&+ \lambda\sum_{\bar{\vec{x}} \in \set{\Omega}_{0}}(1-\langle \nabla \set{O}^i(\vec{x}), \vec{n}^*\rangle)\Big)
\end{align*}

where $\set\Omega$ is the 3D querying space,  $\set\Omega_{0}$ is the zero-level surfaces in the querying space, $\vec{s}^*$ is the ground truth signed-distance, $\vec{n}^*$ is the ground truth normal, and $\nabla$ denotes the spatial gradient. 
We clamp off-surface signed-distances with parameter $\delta$ to concentrate network capacity on details near the surface, as demonstrated in \cite{park2019deepsdf}.For latent space regularization, we impose a Gaussian prior on the object code $L_{latent}(\vec{\alpha}) =\sum_{i=1}^N \| \frac{1}{l} \vec{\alpha}^i \|_2$ and weights of the network  $L_{hyper}(\set{\theta}_o) =\sum_{i=1}^N \frac{1}{l_o} \| \set{\theta}_{o}^i \|_2$ where $l_o$ is the length of $\set{\theta}_{o}$. 
After the pre-training, we fix the object codes and object module weights. 


\subsection{Deformed Object Representation} \label{sec. Deformed Object Representation}
In this section, we extend our nominal shape representation to capture object deformations due to external contacts. \accro~uses two main components to model deformations: First, it summarizes the contact formation (external contact locations and the reaction force) using the Force Module. Second, it uses this summary to  predict a deformation field using the Deformation Module, Fig.~\ref{fig:arch}. Intuitively, the deformation field maps the deformed object back into its nominal shape. In the following, we describe the details of this architecture.

\begin{figure}[thpb]
    \begin{center}
    \includegraphics[width=0.5\textwidth]{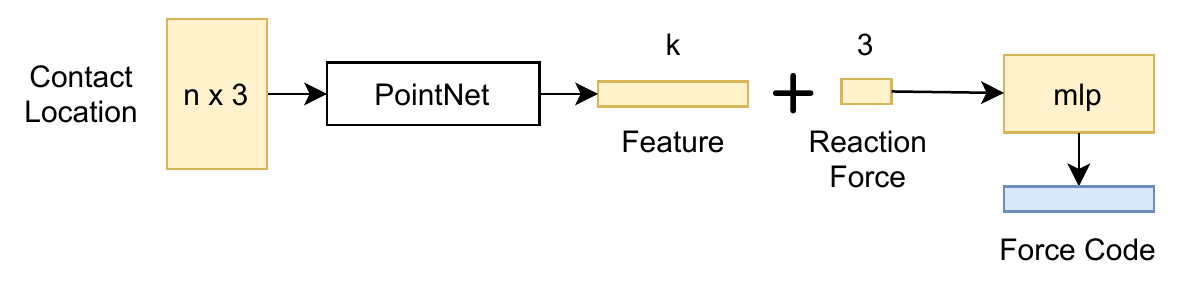}
      \caption{\label{fig:force module} Force Module Architecture}
    \includegraphics[width=0.5\textwidth]{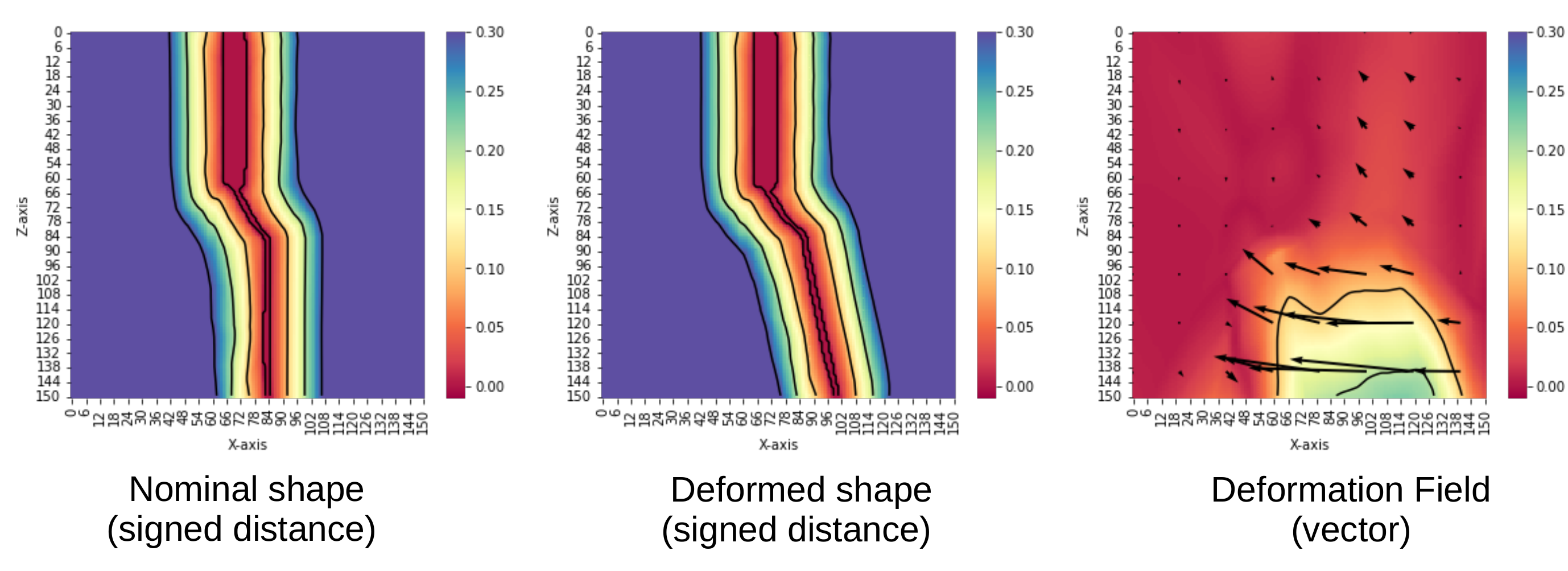} 
  \end{center}
  \caption{Cross-section of x-z plane of the 3D space. The number in the color bar indicates signed distances(left, middle) or the norm of point-wise deformation(right). Arrow indicates the deformation produced for mapping a deformed object's space to the nominal object's space} \label{fig:cross section}  \vspace{-5mm}
\end{figure}

\textbf{Encoding Contact Formations as Force Codes:} The Force Module $\set{F}$ is an encoder that summarizes the contact locations and reaction force ($\set{Q}, \vec{u}$) into a force code $\vec{z} = \vec{F}(\vec{Q}, \vec{u})$. We assume that the contact location set $\set{Q}$ is given as a subset of the nominal point cloud ($\set{Q} \subset{\set{P}}$) and the reaction force  $\vec{u} \in \realnum^3$ is directly measured at the robot's wrist. Point clouds, including the contact location set $\set{Q}$, are unordered and variable in length. In order to effectively summarize $\set{Q}$ into a fixed-length embedding, we utilize the architecture visualized in Fig. \ref{fig:force module}. Our contact location encoder utilizes the PointNet architecture \cite{qi2017pointnet} and has a number of favorable properties including permutation invariance, handling variable length inputs, and is especially effective for capturing low-level features \cite{liu2020morphing}.

\textbf{Deformation Module and Deformation Field:} We define the deformation field as a 3D vector field that pushes a deformed object back to its original (nominal) shape. We illustrate this idea in Fig.~\ref{fig:cross section}, where the left and middle panels depict the signed-distance fields of a nominal and deformed spatula. \accro~recovers the signed-distance field of the nominal shape by adding the deformation field to the SDF of the deformed shape (right panels of Fig.~\ref{fig:cross section}). We highlight that \accro~has learned to focus the deformation field around the boundary of the object with magnitude reflecting the amount of deformation. 

\accro~represents the deformation field as $\vec{D} \big( \vec{x}|\set{\Psi}_{d}(\vec{z},\vec{\alpha}) \big)$, where the deformation module $\set D$ shares the same structure as the object module $\set O$ with parameters $\set\theta_d$ predicted by the hyper-network $\set\Psi_d$. We highlight that $\set\theta_d$ is conditioned on the latent code pair $(\vec{z}, \vec{\alpha}) \in \realnum^{l+m}$ to capture the underlying object-specific deformation behavior. This results in $\set D$ predicting different deformation fields for different objects despite similar contact locations and reaction force measurements. This is desirable because objects may be geometrically similar but deform differently due to varying material properties. We will demonstrate examples of this in Sec.~\ref{sec: Representing Known Deformable Shape}.

To learn the full model, we train the deformation, force, and pre-trained object modules end-to-end using the loss function:
\begin{equation}
    \label{eq:deform_loss}
    L_{deformed} = f_c(\{ \vec x| \vec x \in \set \Omega_0 \})+ \lambda_1 L_{sdf} + \lambda_3 L_{hyper} + \lambda_4 L_{latent}. 
\end{equation} 

The first term solves the optimization in Eq.~\ref{eq:deformation-field} with  on-surface points. Using the shorthand $\vec D_{\set\Psi_d}(\vec x) = \vec{D} \big( \vec{x}|\set{\Psi}_{d}(\vec{z},\vec{\alpha}) \big)$, we can relate the signed-distance fields of the deformed and nominal object via
$s = \text{SDF}(\vec x) = \set O_{\set\Phi_o} \big( \vec x + \set D_{\set\Psi_d}(\vec x) \big)$. We note that the deformed point cloud $\vec p \in \set P$ satisfies $\text{SDF}(\vec p) = \set O_{\set\Phi_o} \big( \vec p + \set D_{\set\Psi_d}(\vec p) \big) = 0$. To learn this mapping, we solve the optimization problem $\underset{\set\theta_d}{\mathrm{argmin}} f_c(\set P)$, where 
\begin{align}
\label{eq:deformation-field}
f_c(\set P) =
\Big(~\text{CD}(\set P + \set D_{\set\Psi_d}(\vec P), \bar{\set{P}}^*)
+  \lambda_{c}\frac{1}{\bar{\vec P}}\sum_{\vec p \in \set P}\| \set D_{\set\Psi_d}(\vec p) \|_2 \Big).
\end{align}
$\bar{\vec P}^*$ is the true nominal point cloud of the same length as $\vec P$, $\lambda_{c}$ is a weighting for the minimal correction prior similar to \cite{DIF} and CD is the Chamfer Distance measured between two point clouds:
\begin{align}
CD(\set{S_1}, \set{S_2})= \frac{1}{|\set{S_1}|}\sum_{\vec{x}\in \set{S_1}} {\min_{\vec{y} \in \set{S_2}}||\vec{x}-\vec{y}||^2_2} + \frac{1}{|\set{S_2}|}\sum_{{\vec{x} \in \set{S_1}}} {\min_{\vec{y}\in \set{S_2}} ||\vec{x}-\vec{y}||^2_2},
\end{align}
Solving Eq.~\ref{eq:deformation-field} with dense point clouds $\bar{\vec P}^*$ and $\set P$ encourages a 1:1 matching between the two while minimizing corrections. This encourages the correspondence shown in Fig.~\ref{fig:correspondence}, allowing $\set\Psi_d(\vec p)$ to be interpreted as physically meaningful deformations or deflections. 

The second term of Eq.~\ref{eq:deform_loss} couples the deformation field to the signed-distance field of the object module: \vspace{-5pt} 
\begin{align} \label{eq: deform_loss_sdf}
L_{sdf} =\sum_{i=1}^M\Big( 
&\sum_{\vec{x} \in \set{\Omega}}| clamp( \set O_{\set\Phi_o} \big( \vec x + \set D_{\set\Psi_d}(\vec x), \delta \big) - clamp(s^*, \delta )| \nonumber \\
&+  \sum_{\vec{x} \in \set{\Omega}_{0}}\lambda_n(1-\langle \nabla \set{O}^{i}(\vec x), \vec{n}^*\rangle)\Big)
\end{align}

where $M$ is the total number of deformed objects. The $L_{hyper}(\vec x)$ and $L_{latent}(\vec x)$ are the Gaussian prior on the latent code and the network parameters. Since we are updating two hyper-networks, $L_{hyper}$ is the weighted sum of $\set \Psi_o$ and $\set \Psi_d $ and the $L_{latent}$ is $\sum_{i=1}^N \| \frac{1}{l} \vec{z} \|_2$.

%% file: text/04-results.tex
\section{Experiments and Results} 
 \label{sec. 4 Experiments and Results}

\subsection{Data Preparation}
To the best of our knowledge, there is no open-sourced synthetic 3D shape data-set (e.g. similar to ShapeNet) containing deformable tools generated by Finite Element Analysis (FEA). To build such a data-set, we start by collecting life-sized 3D models of elastic deformable tools from open-sourced 3D model repository. To simulate deformations for each tool, we utilized MATLAB PDE toolbox, which not only exports nodal analysis in python-friendly extensions, but also provides streamlined automation and high software stability compare to other GUI-based software (CAD, Solidworks). To generate deformations using MATLAB PDE toolbox, we explicitly define fixed constraints, structural boundary loads, and material properties. All constraints and frames are defined with respect to the objects center of mass assuming uniform density. For the analysis, the raw meshes are pre-processed into oriented watertight forms. The simulation output is the nodal analysis of the structural model, containing reaction force ($\vec{u}$) at the fixture, vertexes where the structural boundary loads are applied ($\set{Q}$), and deformed object point clouds ($\set{P}$). We discuss insights for real-world data collection in Sec.~\ref{sec:discussion}.

For the training, we start by normalizing the point cloud with the geometric center at $[0,0,0]$. Next, we sample 25k spatial points in the bounding box with higher density near the surface of the object. For each sample, we calculate ground-truth signed distances by finding the closest point (Euclidean) to the object point cloud. In total, we generated 6 objects, each with 24 unique boundary conditions.
\vspace{-5pt}

\begin{table}
    
        \begin{tabular}{c|c|c|c|c|c|c } 
        \hline
        Metric & \multirow{2}{*}{$CD$} & \multirow{2}{*}{$CD_{m^2}$} &\multicolumn{2}{c|}{$L1$}&\multicolumn{2}{c}{$L1_{m}$}\\
        \cline{4-7}
        (x$10^3$)&&& $\Omega_0$   & $\Omega \setminus \Omega_0$ & $\Omega_0$  & $\Omega \setminus \Omega_0$ \\
        \hline \hline
         Train.Nom.    & 0.7699 & 0.0245   & 0.0026 & 1.9391 & 0.0003 & 0.2493 \\
        \hline 
          Train.Def. & 0.6668 & 0.0220 & 0.0212 & 1.4225 & 0.0028 & 0.1891 \\
        \hline
        Test.Def. & 1.1489 &0.0394 & 0.02578 & 2.1392 & 0.0035 & 0.2859\\
        \hline
        \end{tabular}

    \caption{\textbf{Learning Accuracy :}  \label{tab:Model Accuracy} (Lower is better) Average reconstruction accuracy of deformed objects multiplied by $10^3$. Chamfer Distance ($CD$) and ($CD_{m^2}$) are measured with the reconstructed point cloud (5.6k points) and on-surface points (5.6k points for Train, 0.8k points for Test). $L1$ and $L1_{m}$ distance are evaluated for both on-surface and off-surface points (15.6k $\sim$ 24k points for train, 12k for test). Nom. refers to the nominal shape, and Def. refers to all 144 deformations (6 objects $\times$ 24 deformations). 
}
\vspace{-6mm}
\end{table}

\subsection{Representing Known Deformable Shapes \label{sec: Representing Known Deformable Shape}}

We first evaluate the models ability to represent shapes seen during training. Fig. \ref{fig:deformations} shows the reconstruction of nominal and deformed object geometries. We use the Marching Cube algorithm \cite{lorensen1987marching} to estimate object zero-level sets. We emphasize that only one neural network model was used for the entire data-set. Fig. \ref{fig:deformations} illustrates that the model can accurately fit surface curves and details at sharp edges. While the contact locations $\set Q$ are a subset of nominal point clouds (purple), we mark them on the deformed object (blue) for the clarity. Fig. \ref{fig:deformations} also shows that the force module outputs unique deformation field when given the same contact locations but given different reaction forces.

The performance of the model is quantitatively analyzed in Table \ref{tab:Model Accuracy}. For the evaluation on nominal shape regression, we replace $\set D_{\set \Psi_d}(\vec p)$ with 0, and compare the result with the ground truth  nominal point cloud. The metrics CD and L1 are computed from normalized point clouds while $\text{CD}_{m^2}$ and $\text{L1}_{m}$ are computed at the physical scale of objects. CD is measured with reconstructions of resolution $400^3$ \cite{Gkioxari_2019_ICCV}. Intuitively, the L1 loss is providing the average absolute error between the true and estimated SDF and CD is providing the mean square error loss between the estimated and true reconstructed object geometries using a naive closest point heuristic. We provide both normalized and to scale losses to account for possible scaling effects.   

\begin{figure}[t]
    \centering
    \includegraphics[scale=0.23]{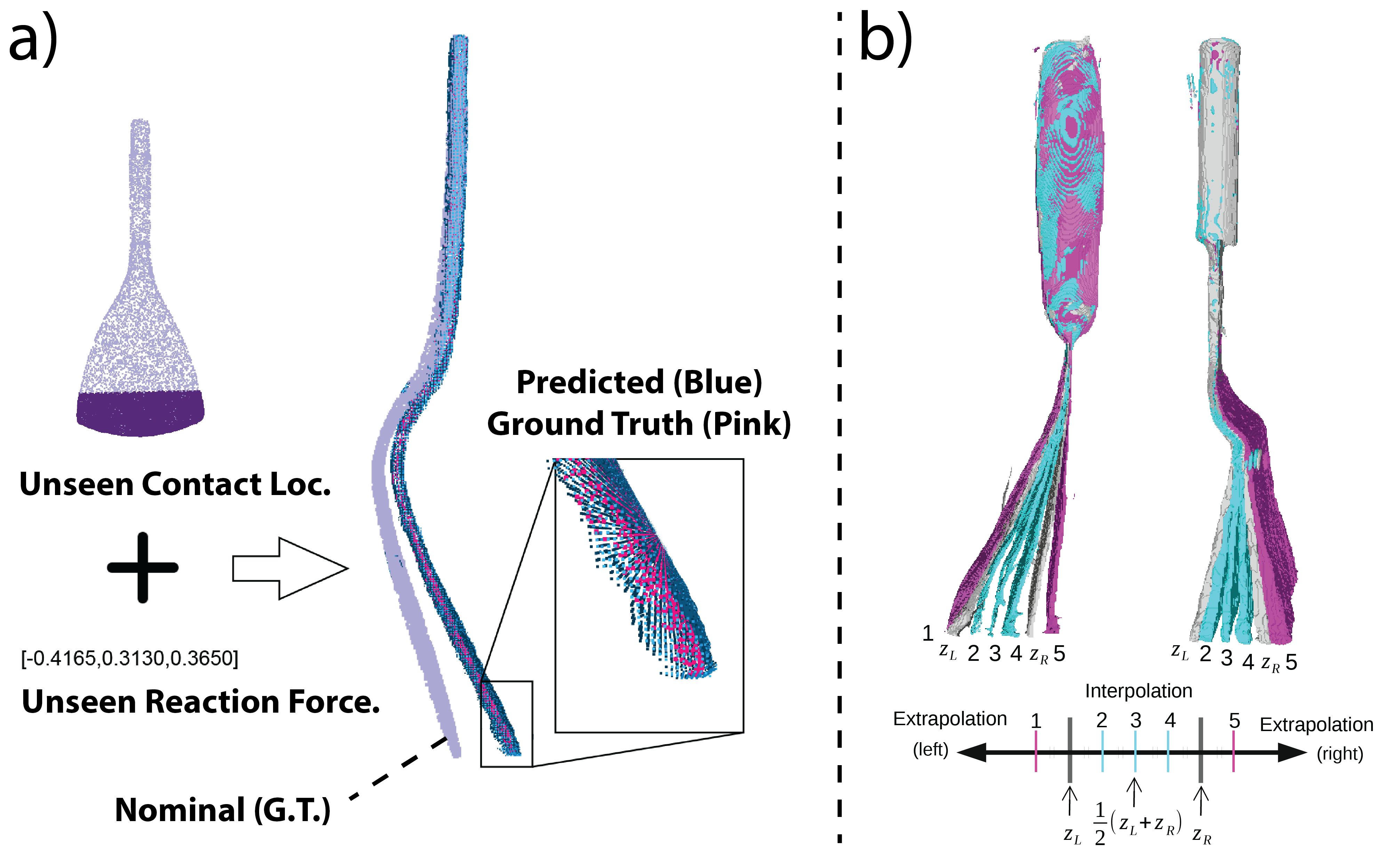}
    \caption{\textbf{a) Generalization:} Example of shape estimation given unseen contact formation during training. \textbf{b) Latent Code Interpolation and Extrapolation:} Two trained force codes $\vec{z}_l$ and $\vec{z}_r$ are interpolated and extrapolated evenly as shown where 1-5 indicate tested force code and corresponding reconstruction result.} 
    \label{fig:gen-latent}
\vspace{-4mm}
\end{figure}

\subsection{Deformation Field Inference \label{sec: deformation field inference}} 
\begin{figure}
    \centering
    \includegraphics[scale=0.6]{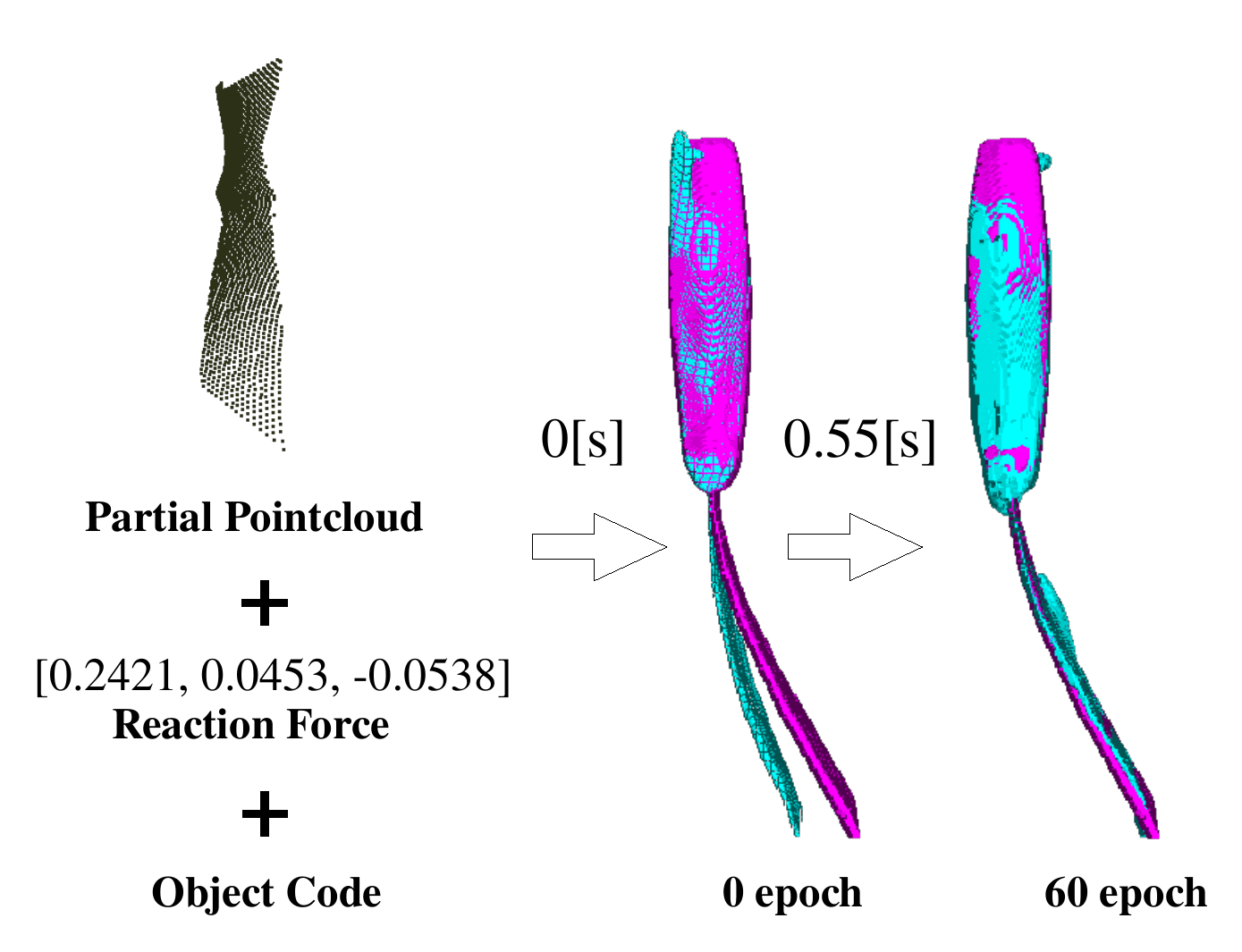}
    \caption{\textbf{Inference:} Reconstructions with inferred deformation field (cyan), ground truth deformed object (magenta)} 
    \label{fig: inference}
\vspace{-4mm}
\end{figure}

We test the model's ability to infer a deformation field given reaction force, partial pointcloud, and object code, seen from the training. Here, we infer the contact feature in Fig.~\ref{fig:force module} to estimate deformation. First,  we randomly initialize the contact feature from $N(0, 0.01^2)$. Then, we update the feature with an L1 loss which only consumes a partial zero-level set: $L_{infer} = \sum_{\vec{x} \in \set{\Omega_o}} |clamp(\set{O}(\vec{x}), \delta)|$. The loss encourages \accro~ to update the contact feature by minimizing the mismatch between the initial guess and the partial observation. Fig. \ref{fig: inference} is a partial pointcloud where the handle and the tip are occluded, rendered in simulation with a single pinhole camera. At epoch 0, the model already makes deformation field fairly close to the ground truth. This shows  \accro's ability to perform state estimation when the vision is missing. As the gradient descent progresses on the contact feature, the estimated deformation converges towards the ground truth. We note that only the on-surface points are used for this experiment, since an RGBD camera would feasibly only give on-surface points in real-world experiments; however, it is also possible boost the inference performance by collecting off-surface samples along the camera ray and augment the partial observation.

\subsection{Generalization and Code Interpolation/Extrapolation}
In this section, we evaluate \accro's ~ability to generalize to unseen contact formations. This functionality is important for robotic applications given the wide variety of contact interactions. To this end, we use the model from Sec.~\ref{sec: Representing Known Deformable Shape} (trained on 144 deformations) and evaluate the reconstruction accuracy of 6 unseen contact formations for the object depicted in Fig.\ref{fig:gen-latent}a). 
To better understand the generalization properties of \accro, we interpolate and extrapolate in latent force code space. This task evaluates the continuity and semantic meaning of the latent space; e.g. whether the reconstructions are similar to direct inter/extrapolation of two deformations.
Given an object, we pick $\vec{z}_l$ and $\vec{z}_r$ among the successfully trained force codes; $\vec{z}_l$ bears the maximum deflection to the left (or -$x$ direction) and $\vec{z}_r$ generates moderate deflections in a random direction. We then linearly interpolate and extrapolate the 32 dimensional force code. Fig.\ref{fig:gen-latent}b) shows the resulting reconstructions. We note the interpolations and extrapolations are smooth, even, and continuous. This suggests a well-formed latent force space and explains the effectiveness of the model in generalizing to unseen contact formations. 

We note that the reconstruction for extrapolation (1) is omitted from the object on the right in Fig.~\ref{fig:gen-latent}. This is because of a failure case resulting in a noisy and poor quality reconstruction. We found that the extrapolation results can be inaccurate if it exceeds the bound of maximum deflections seen in the training data. 


\subsection{Dense Correspondence}
\begin{figure}
    \centering
    \includegraphics[width=0.5\textwidth]{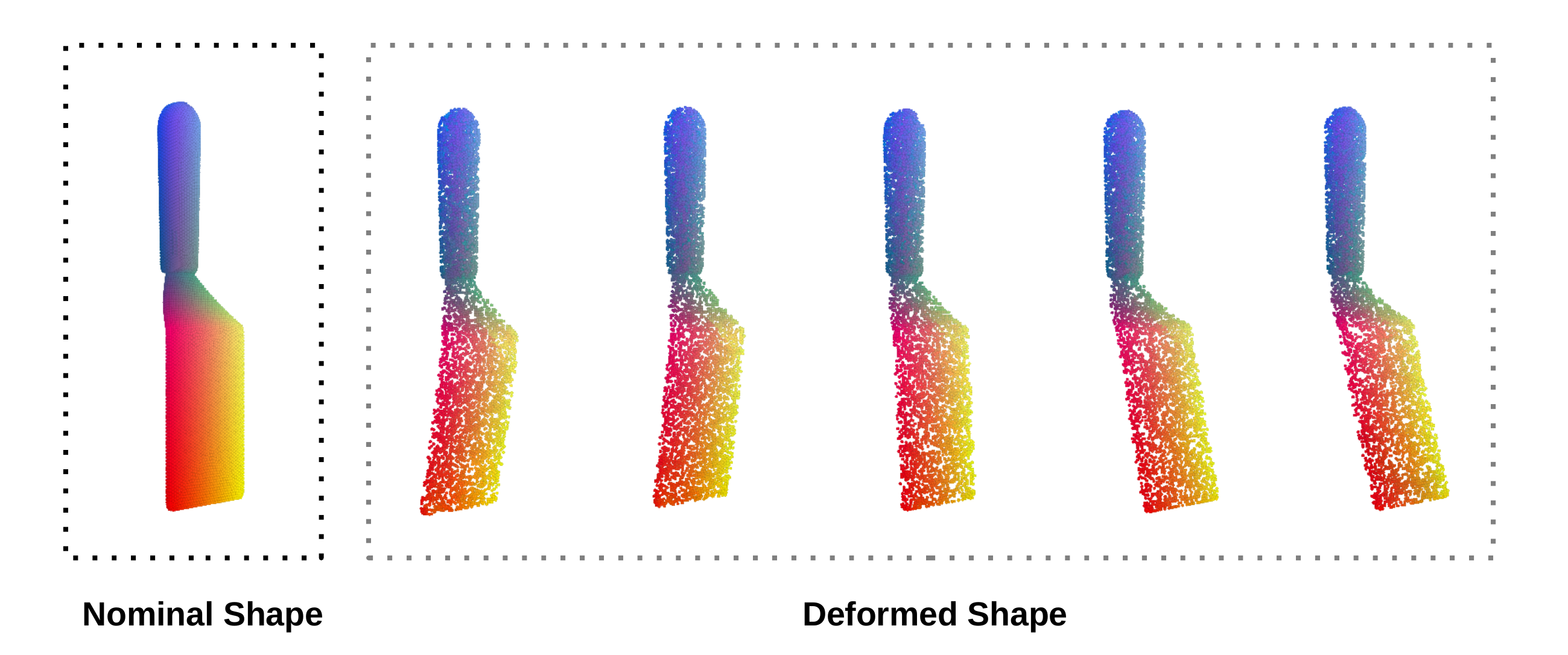}
  \caption{Point correspondences are maintained under the deformation field w/o explicit supervision.}
  \label{fig:correspondence} \vspace{-4mm}
\end{figure}

An important feature of \accro~is its ability to infer dense correspondence without explicit supervision, as shown in Fig.~\ref{fig:correspondence}. Maintaining dense correspondence is important for tracking during robotic task execution. To derive Fig.~\ref{fig:correspondence}, we ``paint'' the nominal shape, then apply a variety of contact formations and track the painted points. The paintwork shows the consistency of deformation of points on the object. Correspondence for deformable objects is useful for tracking not only the pose of the object, but also the relative deflection of points. To illustrate, we can estimate the deflection of any point by subtracting its initial position from its last using correspondences: $\Delta \vec p = \set D_{\set\Psi_d}(\vec p_{last}) - \set D_{\set\Psi_d}(\vec p_{init})$. \\



%% file: text/02-related-work.tex
\vspace{-10pt}
\section{Related Work}
Deformable object manipulation has a long history in robotics. To date, two general approaches have been attempted: model-based and data-driven. In the following, we focus the related work on semi-deformable and elastically deformable techniques.

\textbf{Elastic Object Modeling for Robotic Manipulation:} The current representations we use in robotics are, by-and-large, borrowed from continuum mechanics. Two predominant examples are Finite Element Models (FEMs) \cite{essa1992unified,ficuciello2018fem,frank2014learning, muller2004interactive,sengupta2019tracking} and particle models \cite{howard2000intelligent,nurnberger1998problem,tonnesen2000dynamically}. FEMs numerically solve coupled constituents equations defined over object meshes for object deformations as a function of applied boundary conditions. FEM methods are typically prohibitively computationally expensive to run online and require careful system identification \cite{arriola2020modeling,jangir2020dynamic,sun2008learning}. Particle models approximate objects as a set of particles related to each other by constitutive laws. One disadvantage of particle systems is that the object surface is not explicitly defined. Therefore, maintaining the initial shape of the deforming object is difficult, introducing issues in tracking the return of elastic objects to their original shape after deformation during manipulation \cite{sanchez2018robotic,arriola2020modeling}. More recently, learning methods have been integrated to assist in nonlinear modeling of deformable object \cite{zhang2019neural, arriola2017multimodal}. The computational costs of model-based approaches and the challenges with integration with robotic sensing modalities (RGB-D cameras and F/T sensors) pose significant challenges to these methods.

\textbf{Learned representations of 3D geometries:} Learned implicit representations of complex 3D geometries have recently gained traction in the computer vision and graphics communities \cite{park2019deepsdf,pratt2017fcnn,sitzmann2020implicit,tancik2020fourier,mescheder2019occupancy,liu2020morphing,DIF,kpconv}. These approaches map low-dimensional 2D (e.g. image coordinates) or 3D (e.g. spatial coordinates) points to a variety of geometric representations including signed-distances, occupancy, volume density \cite{chen2019learning,deng2019nasa,genova2019learning,genova2020local,michalkiewicz2019implicit} and Neural radiance fields that learn to represent a scene’s structure from posed RGB images (3D shape representations to be learned using only 2D images as supervision \cite{mildenhall2020nerf}. These approaches are promising for robotic applications for three primary reasons. First, they are computationally efficient owing to their neural network representations. Second, they directly operate on images and RGB-D data-streams that are commonly available in robotics applications. Third, they produce high-fidelity object representations. However, in addition to the integration of tactile feedback and deformation prediction, advances need to be made in state-estimation, planning, and controls.

%% file: text/05-discussion.tex
\section{Discussion \& Limitations}\label{sec:discussion}

\textbf{Summary:} The fundamental principle driving \accro~is the ability to learn deformation fields informed by visio-tactile sensing. \accro~is the first learned implicit method to integrate tactile and visual feedback while modeling object deformations subject to external contacts. The representation has arbitrary resolution and is cheap to evaluate for point-wise sampling. Additionally, the latent code is well-behaved and can be used for inference.

\textbf{Representation Choice: \label{appendix: representation choice}} 
In this paper, we chose signed distance fields as an instance of continuous neural representations. This is motivated by the SOTA shape representations results from the computer vision community \cite{park2019deepsdf,sitzmann2020implicit,tancik2020fourier}. These continuous representations offer a number of advantages over their discrete counterparts (e.g. meshes and voxel grids) including greater fidelity, flexibility, and compactness~\cite{park2019deepsdf,sitzmann2020implicit}. Compared to meshes, continuous neural representations are particularly good at handling topology changes \cite{chen2019learning}. For deformable objects, this could either be topology changes between different instances of the category, or an object undergoing a topology change due to external forces. Although our experiments do not probe this property explicitly, it does lay the foundation for future work to study these properties. 

We are further motivated by real-world hardware designs: RGB-D cameras collect point clouds that are measurements of the zero-level sets of object geometries. This allows point clouds to be naturally treated by the signed distance function without additional modification. In addition, our representation directly uses reaction forces at the wrist. These functionalities significantly simplify the state-estimation problem and allow for a more direct integration of the representation. 

While it is possible to apply the idea of deformation field learning to explicit or volumetric methods, we must resort to meshing, voxelization, and a number of other post processing steps that our method does not require. These steps typically require significant engineering and domain knowledge that may not be available a priori. Further, the perception model converting visio-tactile feedback to meshes is often complex and requires careful tuning.

\textbf{Towards Real-World Robotic Applications: \label{appendix: Towards Real-World Robotic Applications}} In this paper, we demonstrated the proof of concept of our method assuming no occlusions and accurate contact patch estimation -- both are non-trivial perception problems in real-world applications. It is possible to collect occlusion-free zero-level sets of nominal shapes through turntables/multi-perspective camera setups (e.g., hanging the object using strings) or utilizing high-resolution tactile sensors such as \cite{donlon2018gelslim,alspach2019soft} that provide the imprints of the object in the grasp that can be combined with external vision. However, occlusions are inevitable during contact. One potential approach is to make contact with known and fixed external features such that the contact patch can be closely approximated.

An alternative approach is to forgo the PointNet architecture (Fig.~\ref{fig:force module}) entirely and learn the contact feature for each contact formation. This approach no longer explicitly reasons over the contact patch but can still be used effectively for deformation estimation -- as demonstrated by inferring contact code and deformation fields from partial views with no explicitly contacts. An interesting extension of \accro~ would be the addition of a contact feature decoder that directly infers contact location from the converged contact feature from Sec.~\ref{sec: deformation field inference}. Our inference results and the current iteration of \accro~leave out object pose estimation and sensor uncertainty which simplifies both training and inference. Our next goal is to jointly infer the object pose and deformation geometry given real-world (occlusion and uncertain) data.

\textbf{Dynamics:} \accro~enables high-quality reconstruction of static scenes, as is common with the majority of 3D learned implicit representations. The rich and continuous learned  latent code opens the door to including dynamic (temporal) deformations; however, learning a single embedding over both geometry and time remains a major challenge that we look forward to tackling.